\pdfoutput=1
\documentclass{article} 
\usepackage{iclr2020_conference,times}
\usepackage{times}
\usepackage{epsfig}
\usepackage{graphicx}
\usepackage{amsmath}
\usepackage{amssymb}
\usepackage{amsthm}
\usepackage{caption}
\usepackage{subcaption}

\usepackage{authblk}

\theoremstyle{proposition}
\newtheorem{proposition}{Proposition}[section]
\theoremstyle{definition}
\newtheorem{definition}{Definition}[section]
\theoremstyle{definition}
\newtheorem{remark}{Remark}
\DeclareMathOperator{\argmax}{arg\,max}

\usepackage{hyperref}
\usepackage{url}

\title{Information Plane Analysis of Deep Neural Networks via Matrix--Based R\'enyi's Entropy and Tensor Kernels}

\author[, 1]{Kristoffer Wickstr\o m\thanks{ Corresponding author: kristoffer.k.wickstrom@uit.no. Wickstr\o m, L\o kse, Kampffmeyer and Jenssen are all at the UiT Machine Learning Group (https://machine-learning.uit.no/)}}
\author[1]{Sigurd L\o kse}
\author[1]{Michael Kampffmeyer}
\author[2]{Shujian Yu}
\author[2]{Jose Principe}
\author[1]{Robert Jenssen}
\affil[1]{Department of Physics and Technology, UiT The Arctic University of Norway}
\affil[2]{Computational NeuroEngineering Laboratory, Department of Electrical and Computer Engineering, University of Florida}

\iclrfinalcopy 

\newcommand{\vectorize}[1]{\ensuremath{\mathbf{#1}}}
\newcommand{\K}{\vectorize{K}}
\newcommand{\A}{\vectorize{A}}

\newcommand{\x}{\vectorize{x}}
\newcommand{\y}{\vectorize{y}}

\DeclareMathOperator{\tr}{tr}

\begin{document}

\maketitle

\setlength{\textfloatsep}{8pt plus 0pt minus 7pt}
\begin{abstract}
   Analyzing deep neural networks (DNNs) via information plane (IP) theory has gained tremendous attention recently as a tool to gain insight into, among others, their generalization ability. However, it is by no means obvious how to estimate mutual information (MI) between each hidden layer and the input/desired output, to construct the IP. For instance, hidden layers with many neurons require MI estimators with robustness towards the high dimensionality associated with such layers. MI estimators should also be able to naturally handle convolutional layers, while at the same time being computationally tractable to scale to large networks. None of the existing IP methods to date have been able to study truly deep Convolutional Neural Networks (CNNs), such as the e.g.\ VGG-16. In this paper, we propose an IP analysis using the new matrix--based R\'enyi's entropy coupled with tensor kernels over convolutional layers, leveraging the power of kernel methods to represent properties of the probability distribution independently of the dimensionality of the data. The obtained results shed new light on the previous literature concerning small-scale DNNs, however using a completely new approach. Importantly, the new framework enables us to provide the first comprehensive IP analysis of contemporary large-scale DNNs and CNNs, investigating the different training phases and providing new insights into the training dynamics of large-scale neural networks.
\end{abstract}

\section{Introduction}

\begin{figure*}[h]
    \centering
    \includegraphics[trim={0 0 0 0}, clip, width=\textwidth]{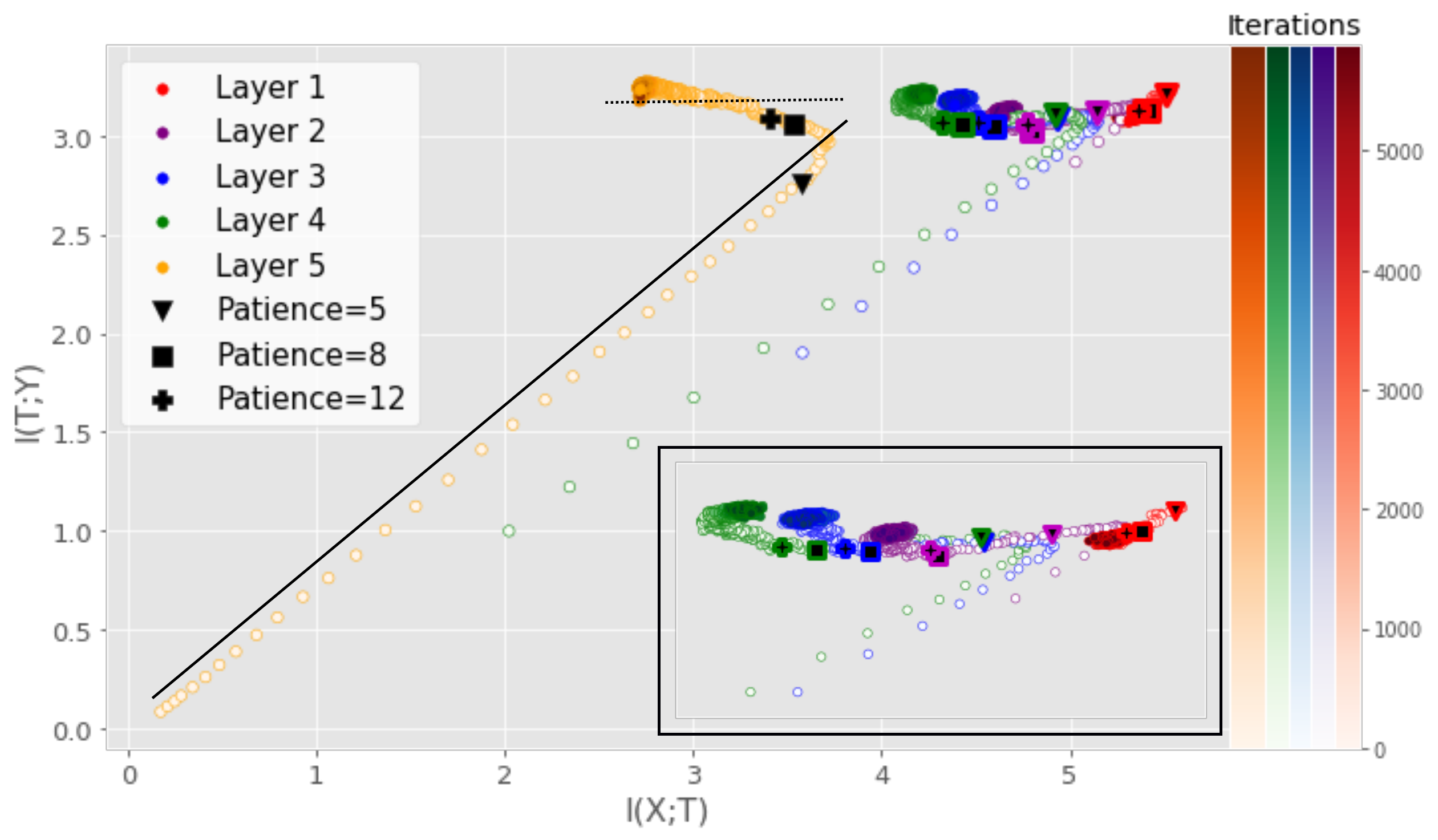}
    \caption{IP obtained using our proposed estimator for a small DNN averaged over 5 training runs. The solid black line illustrates the fitting phase while the dotted black line illustrates the compression phase. The iterations at which early stopping would be performed assuming a given patience parameter are highlighted. Here, patience denotes the number of iterations that need to pass without progress on a validation set before training is stopped to avoid overfitting. It can be observed that for low patience values, training will stop before the compression phase. For the benefit of the reader, the bottom right corner displays a magnified version of the first four layers.}
    \label{fig:front_page}
\end{figure*}

Although Deep Neural Networks (DNNs) are at the core of most state--of--the art systems in computer vision, the theoretical understanding of such networks is still not at a satisfactory level \citep{DBLP:journals/corr/Shwartz-ZivT17}. In order to provide insight into the inner workings of DNNs, the prospect of utilizing the Mutual Information (MI), a measure of dependency between two random variables, has recently garnered a significant amount of attention~\citep{Cheng_2018_ECCV, 8683351, michael2018on, DBLP:journals/corr/Shwartz-ZivT17, DBLP:journals/corr/abs-1804-06537, DBLP:journals/corr/abs-1804-00057}. Given the input variable $X$ and the desired output $Y$ for a supervised learning task, a DNN is viewed as a transformation of $X$ into a representation that is favorable for obtaining a good prediction of $Y$. By treating the output of each hidden layer as a random variable $T$, one can model the MI $I(X;T)$ between $X$ and $T$. Likewise, the MI $I(T;Y)$ between $T$ and $Y$ can be modeled. The quantities $I(X;T)$ and $I(T;Y)$ span what is referred to as the Information Plane (IP). Several works have demonstrated that one may unveil interesting properties of the training dynamics by analyzing DNNs in the form of the IP. Figure~\ref{fig:front_page}, produced using our proposed estimator, illustrates one such insight that is similar to the observations of \cite{DBLP:journals/corr/Shwartz-ZivT17}, where training can be separated into two distinct phases, the fitting phase and the compression phase. This claim has been highly debated as subsequent research has linked the compression phase to saturation of neurons~\citep{michael2018on} or clustering of the hidden representations~\citep{DBLP:journals/corr/abs-1810-05728}.

{\bf Contributions}\hspace{0.2cm} We propose a novel approach for estimating MI, wherein a kernel tensor-based estimator of R\'enyi's entropy allows us to provide the first analysis of large-scale DNNs as commonly found in state-of-the-art methods. We further highlight that the multivariate matrix--based approach, proposed by \cite{DBLP:journals/corr/abs-1808-07912}, can be viewed as a special case of our approach. However, our proposed method alleviates numerical instabilities associated with the multivariate matrix--based approach, which enables estimation of entropy for high-dimensional multivariate data. Further, using the proposed estimator, we investigate the claim of \cite{Cheng_2018_ECCV} that the entropy $H(X) \approx I(T;X)$ and $H(Y) \approx I(T;Y)$ in high dimensions (in which case MI-based analysis would be meaningless) and illustrate that this does not hold for our estimator. Finally, our results indicate that the compression phase is apparent mostly for the training data, particularly for more challenging datasets. By utilizing a technique such as early-stopping, a common technique to avoid overfitting, training tends to stop before the compression phase occurs (see Figure~\ref{fig:front_page}). This may indicate that the compression phase is linked to the overfitting phenomena.

\section{Related Work}
\label{sec:related}
Analyzing DNNs in the IP was first proposed by \cite{7133169} and later demonstrated by \cite{DBLP:journals/corr/Shwartz-ZivT17}. Among other results, the authors studied the evolution of the IP during the training process of DNNs and noted that the process was composed of two different phases. First, an initial fitting phase where $I(T;Y)$ increases, followed by a phase of compression where $I(X;T)$ decreases. These results were later questioned by \cite{michael2018on}, who argued that the compression phase is not a general property of the DNN training process, but rather an effect of different activation functions. However, a recent study by \cite{8683351} seems to support the claim of a compression phase, regardless of the activation function. The authors argue that the base estimator of MI utilized in~\cite{michael2018on} might not be accurate enough and demonstrate that a compression phase does occur, but the amount of compression can vary between different activation functions. Another recent study by \citep{chelombiev2018adaptive} also reported a compression phase, but highlighted the importance of adaptive MI estimators. They also showed that when L2-regularization was included in the training compression was observed, regardless of activation function. Also, some recent studies have discussed the limitations of the IP framework for analysis and optimization for particular types of DNN (\cite{kolchinsky2018caveats, 8680020}).

On a different note, \cite{Cheng_2018_ECCV} proposed an evaluation framework for DNNs based on the IP and demonstrated that MI can be used to infer the capability of DNNs to recognize objects for an image classification task. Furthermore, the authors argue that when the number of neurons in a hidden layer grows large, $I(T;X)$ and $I(Y;T)$ barely change and are approximately deterministic, i.e. $I(T;X)\approx H(X)$ and $I(T;Y)\approx H(Y)$. Therefore, they only model the MI between $X$ and the last hidden layer, that is the output of the network, and the last hidden layer and $Y$. 

\cite{DBLP:journals/corr/abs-1804-00057} initially proposed to utilize empirical estimators for R\'enyi's MI for investigating different data processing inequalities in stacked autoencoders (SAEs). They also claimed that the compression phase in the IP of SAEs is determined by the values of the SAE bottleneck layer size and the intrinsic dimensionality of the given data. \cite{DBLP:journals/corr/abs-1808-07912} also extended the empirical estimators for R\'enyi's MI to the multivariate scenario and applied the new estimator to simple but realistic Convolutional Neural Networks (CNNs) \citep{DBLP:journals/corr/abs-1804-06537}. However, the results so far suffer from high computational burden and are hard to generalize to deep and large-scale CNNs.

\section{Matrix--Based Mutual Information}\label{sec:matrix_renyi}

Here, we review R\'enyi's $\alpha$-order entropy and its multivariate extension proposed by \cite{DBLP:journals/corr/abs-1808-07912}.

\subsection{Matrix--Based R\'enyi's alpha-Order Entropy}

R\'{e}nyi's $\alpha$-order entropy is a generalization of Shannon's entropy~\citep{ Shannon:2001:MTC:584091.584093, rényi1961}. For a random variable $X$ with probability density function (PDF) $f(\mathbf{x})$ over a finite set $\mathcal{X}$, R\'{e}nyi's $\alpha$-order entropy is defined as:
\begin{equation}\label{eq:alpha renyi}
    H_\alpha(f)=\frac{1}{1-\alpha}\log\int_\mathcal{X}f^\alpha (\mathbf{x})d\mathbf{x}
\end{equation}
Equation~\ref{eq:alpha renyi} has been widely applied in machine learning \citep{Principe:2010:ITL:1855180}, and the particular case of $\alpha=2$, combined with Parzen window density estimation \citep{parzen1962}, form the basis for Information Theoretic Learning \citep{Principe:2010:ITL:1855180}.
However, accurately estimating PDFs in high-dimensional data, which is typically the case for DNNs, is a challenging task.
To avoid the problem of high-dimensional PDF estimation, \cite{DBLP:journals/corr/abs-1211-2459} proposed a non-parametric framework for estimating entropy directly from data using infinitely divisible kernels \citep{DBLP:journals/corr/abs-1301-3551} with similar properties as R\'{e}nyi's $\alpha$-order entropy.

\begin{definition}{\citep{DBLP:journals/corr/abs-1211-2459}}
Let $\x_i\in \mathcal{X},\; i = 1, 2, \ldots, N$
    denote data points and let $\kappa: \mathcal{X}\times\mathcal{X}\mapsto\mathbb{R}$ be
    an infinitely divisible positive definite
    kernel \citep{10.2307/27641890}.
Given the kernel matrix $\K \in \mathbb{R}^{N \times N}$ with
    elements $(\K)_{ij} = \kappa(\x_i, \x_j)$ and
    the matrix $\A,\; (\A)_{ij} = \frac{1}{N}
    \frac{(\K)_{ij}}{\sqrt{(\K)_{ii}(\K)_{jj}}}$,
    the matrix--based  R\'enyi's $\alpha$--order entropy is given by
    \begin{equation}\label{eq:def renyi}
        S_{\alpha}(\A) = \frac{1}{1 - \alpha} \log_2 \Big( \tr(\A^\alpha) \Big) = \frac{1}{1 - \alpha} \log_2 \Bigg[ \sum\limits_{i=1}^N \lambda_i(\A)^\alpha \Bigg]
    \end{equation}
    where $\tr(\cdot)$ denotes the trace, and $\lambda_i(\A)$ denotes the $i^{\text{th}}$ eigenvalue of $\A$.

\end{definition}

Not all estimators of entropy have the same property \citep{Paninski:2003:EEM:795523.795524}. The ones developed for Shannon suffer from the curse of dimensionality \citep{1114861}. In contrast, the matrix--based Renyi's entropy shown in Equation \ref{eq:def renyi} have the same functional form of the statistical quantity in a Reproducing Kernel Hilbert Space (RKHS). Essentially, we are projecting marginal distribution to an RKHS to measure entropy and mutual information. This is similar to the approach of maximum mean discrepancy and the kernel mean embedding \citep{Gretton:2012:KTT:2188385.2188410, fuku}. Moreover, \cite{DBLP:journals/corr/abs-1211-2459} showed that under certain conditions the difference between the estimators and the true quantity can be bounded, as shown in Proposition \ref{prop:bound} in Appendix \ref{sec:bound}. Notice that the dimensionality of the data does not appear in Proposition \ref{prop:bound}. This makes the matrix-based entropy estimator more robust for high-dimensional data compared to KNN and KDE based information estimators used in previous works \citep{michael2018on, chelombiev2018adaptive}, which have difficulties handling high-dimensional data \citep{1114861}. Also, there is no need for any binning procedure utilized in previous works \citep{DBLP:journals/corr/Shwartz-ZivT17}, which are known to struggle with the relu activation function commonly used in DNN \citep{michael2018on}.

\begin{remark}
In the limit when $\alpha \rightarrow 1$,
Equation~\ref{eq:def renyi} reduces to the matrix--based von Neumann entropy \citep{nielsen2000quantum} that resembles Shannon's definition over probability states, and can be expressed as
\begin{equation}\label{eq:shannon}
    \lim_{\alpha \rightarrow 1}S_{\alpha}(\A)=-\sum_{i = 1}^N \lambda_i(\A) \log_2[\lambda_i(\A)].
\end{equation}
 For completeness, the proof of Equation~\ref{eq:shannon} can be found in Appendix \ref{sec:appendixA}.
\end{remark}

In addition to the definition of matrix based entropy, \cite{DBLP:journals/corr/abs-1211-2459} define the joint entropy between $\x \in \mathcal{X}$ and $\y \in \mathcal{Y}$ as
    \begin{equation}\label{eq:joint renyi entropy}
        S_{\alpha}(\A_\mathcal{X}, \A_\mathcal{Y}) =
        S_\alpha\Bigg(\frac{\A_\mathcal{X} \circ \A_\mathcal{Y}}{\text{tr}(\A_\mathcal{X} \circ \A_\mathcal{Y})}\Bigg),
    \end{equation} 
    where $\x_i$ and $\y_i$ are two different representations
    of the same object and $\circ$ denotes the Hadamard product.
Finally, the MI is, similar to Shannon's formulation, defined as

\begin{equation}\label{eq:mutual info}
    I_\alpha(\A_\mathcal{X}; \A_\mathcal{Y}) =
    S_\alpha(\A_\mathcal{X}) + S_\alpha(\A_\mathcal{Y})-S_\alpha(\A_\mathcal{X}, \A_\mathcal{Y}).
\end{equation}

\subsection{Multivariate Matrix-Based R\'enyi's alpha-entropy Functionals} \label{sec:multivariate}

The matrix-based R\'enyi's $\alpha$-order entropy functional is not suitable for estimating the amount of information of the features produced by a convolutional layer in a DNN as the output consists of $C$ feature maps, each represented by their own matrix, that characterize different properties of the same sample. \cite{DBLP:journals/corr/abs-1808-07912} proposed a multivariate extension of the matrix--based R\'enyi's $\alpha$-order entropy, which computes the joint-entropy among $C$ variables as
\begin{equation}\label{eq:multivariate_renyi}
    S_\alpha(\A_1,\ldots,\A_C) =
    S_\alpha\Big(\frac{\A_1\circ\ldots\circ\A_C}{\tr(\A_1\circ\ldots\circ\A_C)}\Big),
\end{equation}
where $(\A_1)_{ij}=\kappa_1(\x_i^{(1)}, \x_j^{(1)})$, ..., $(\A_C)_{ij}=\kappa_C(\x_i^{(C)}, \x_j^{(C)})$. \cite{DBLP:journals/corr/abs-1804-06537} also demonstrated how Equation~\ref{eq:multivariate_renyi} could be utilized for analyzing synergy and redundancy of convolutional layers in DNN, but noted that this formulation can encounter difficulties when the number of feature maps increases, such as in more complex CNNs.
Difficulties arise due to the Hadamard products in Equation~\ref{eq:multivariate_renyi}, given that each element of $A_c,\; c \in \{1, 2, \ldots, C\}$, takes on a value between 0 and $\frac{1}{N}$, and the product of $C$ such elements thus tends towards 0 as $C$ grows. \cite{DBLP:journals/corr/abs-1804-06537} reported such challenges when attempting to model the IP of the VGG16 \citep{DBLP:journals/corr/SimonyanZ14a}.

\section{Tensor-Based Mutual Information}\label{sec:tensor_renyi}

To estimate information theoretic quantities of features produced by convolutional layers and address the limitations discussed above we introduce our tensor-based approach for estimating entropy and MI in DNNs, and show that the multivariate approach in section~\ref{sec:multivariate} arises as a special case.
    
\subsection{Tensor Kernels for Mutual Information Estimation}
The output of a convolutional layer is represented as a tensor $\mathbb{X}_i \in \mathbb{R}^{C}\otimes \mathbb{R}^{H} \otimes \mathbb{R}^{W}$ for a data point $i$.
As discussed above, the matrix--based R\'enyi's $\alpha$-entropy estimator can not include tensor data without modifications. To handle the tensor based nature of convolutional layers we propose to utilize tensor kernels \citep{signoretto2011kernel} to produce a kernel matrix for the output of a convolutional layer. A tensor formulation of the radial basis function (RBF) kernel can be stated as
\begin{equation}\label{eq:tensor kernel}
    \kappa_\text{ten}(\mathbb{X}_i , \mathbb{X}_j) = e^{-\frac{1}{\sigma ^2}\|\mathbb{X}_i -\mathbb{X}_j  \|_F^2},
\end{equation}
where $\| \cdot \|_F$ denotes the Frobenius norm and $\sigma$ is the kernel width parameter. In practice, the tensor kernel in Equation~\ref{eq:tensor kernel} can be computed by reshaping the tensor into a vectorized representation while replacing the Frobenius norm with a Euclidean norm. We estimate the MI in Equation~\ref{eq:mutual info} by replacing the matrix $\mathbf{A}$ with
\begin{equation}
\begin{aligned}
    (\A_\text{ten})_{ij} &= 
        \frac{1}{N} \frac{(\K_\text{ten})_{ij}}{\sqrt{(\K_\text{ten})_{ii} (\K_\text{ten})_{jj}}} \\ 
        &= \frac{1}{N} \kappa_\text{ten}(\mathbb{X}_i, \mathbb{X}_j).
\end{aligned}
\end{equation}

\subsection{Tensor-Based Approach Contains Multivariate Approach as Special Case}
Let $\x_i(\ell) \in \mathbb{R}^{HWC}$ denote the vector representation of data point $i$ in layer $\ell$ 
    and let $\x_i^{(c)}(\ell) \in \mathbb{R}^{HW}$ denote its representation produced by filter $c$.
In the following, we omit the layer index for ease of notation, but assume
    it is fixed. Consider the case when $\kappa_c(\cdot, \cdot)$ is an RBF kernel with kernel width
    parameter $\sigma_c$.
That is $\kappa_c(\x_i^{(c)}, \x_j^{(c)}) = e^{-\frac{1}{\sigma_c^2} \|\x_i^{(c)} - \x_j^{(c)}\|^2}$.
In this case, $\A_c = \frac{1}{N} \K_c$ and
\begin{equation*}
\begin{aligned}
    \frac{\A_1 \circ \ldots \circ \A_C}{\tr(\A_1 \circ \ldots \circ \A_C)}
        &= \frac{
                \frac{1}{N} \K_1 \circ \ldots \circ \frac{1}{N} \K_C
            }{
                \tr(\frac{1}{N} \K_1 \circ \ldots \circ \frac{1}{N} \K_C)
            } \\
        &= \frac{1}{N^C} \frac{\K_1 \circ \ldots \circ \K_C}{\frac{1}{N^C}\tr(\K_1 \circ \ldots \circ \K_C)} \\
        &= \frac{1}{N} \K_1 \circ \ldots \circ \K_C,
\end{aligned}
\end{equation*}
    since $\tr(\K_1 \circ \ldots \circ \K_C) = N$.
Thus, element $(i, j)$ is given by
\begin{equation*}
\begin{aligned}
    \left(\frac{\A_1 \circ \ldots \circ \A_C}{\tr(\A_1\circ \ldots \circ \A_C)}\right)_{ij}
        &= \frac{1}{N} \prod_{c = 1}^C (\K_c)_{ij} \\
        &= \frac{1}{N} e^{-\sum_{c = 1}^C \frac{1}{\sigma_c^2} \|\x_i^{(c)} - \x_j^{(c)}\|^2}. 
\end{aligned}
\end{equation*}
If we let $\sigma = \sigma_1 = \sigma_2 = \ldots = \sigma_C$, this expression is reduced to
\begin{equation}
\begin{aligned}
\frac{1}{N} e^{-\frac{1}{\sigma^2} \sum_{c = 1}^C \|\x_i^{(c)} - \x_j^{(c)}\|^2}
    &= \frac{1}{N} e^{-\frac{1}{\sigma^2} \|\x_i - \x_j\|^2} \\ 
    &= \frac{1}{N} \kappa_\text{ten}(\mathbb{X}_i, \mathbb{X}_j).
\end{aligned}
\end{equation}
Accordingly, $S_\alpha(\A_\text{ten}) = S_\alpha(\A_1,\ldots,\A_C)$ implying that 
    the tensor method is equivalent to the multivariate
    matrix--based joint entropy when the width parameter is equal
    within a given layer.
However, the tensor-based approach eliminates the effect
    of numerical instabilities one encounters in
    layers with many filters, thereby enabling training of complex neural networks. 

\subsection{Choosing the Kernel Width} \label{sec:kernel_width}
With methods involving RBF kernels, the choice of the kernel width parameter, $\sigma$, is always critical. For supervised learning problems, one might choose this parameter by cross-validation based on validation accuracy, while in unsupervised problems one might use a rule of thumb \citep{shi2000normalized, shi2009data, silverman1986density}. However, in the case of estimating MI in DNNs, the data is often high dimensional, in which case unsupervised rules of thumb often fail \citep{shi2009data}.

In this work, we choose $\sigma$ based on an optimality criterion.
Intuitively, one can make the following observation: A good kernel matrix should reveal the class structures present in the data.
This can be accomplished by maximizing the so--called \textit{kernel alignment} loss \citep{cristianini2002kernel} between the kernel matrix of a given layer, $\K_\sigma$, and the label kernel matrix, $\K_y$. The kernel alignment loss is defined as
\begin{equation}\label{eq:kernel loss}
    A(\K_a, \K_b) = \frac{\langle \K_a, \K_b \rangle_F}{\|\K_a\|_F \|\K_b\|_F},
\end{equation}
    where $\|\cdot\|_F$ and $\langle \cdot, \cdot \rangle_F$ denotes the Frobenius norm
    and inner product, respectively.
Thus, we choose our optimal $\sigma$ as
\[
    \sigma^* = \argmax_\sigma A(\K_\sigma, \K_y).
\]
To stabilize the $\sigma$ values across mini batches, we employ
    an exponential moving average, such that in layer $\ell$
    at iteration $t$, we have
\[
    \sigma_{\ell, t} = \beta \sigma_{\ell, t - 1} + (1 - \beta) \sigma_{\ell, t}^*,
\]
    where $\beta \in [0, 1]$ and $\sigma_{\ell, 1} = \sigma_{\ell, 1}^*$.

\begin{figure*}[tb]
    \centering
    \includegraphics[trim={0.2cm 0.3cm 0.2cm 0.2cm}, clip, width=\linewidth]{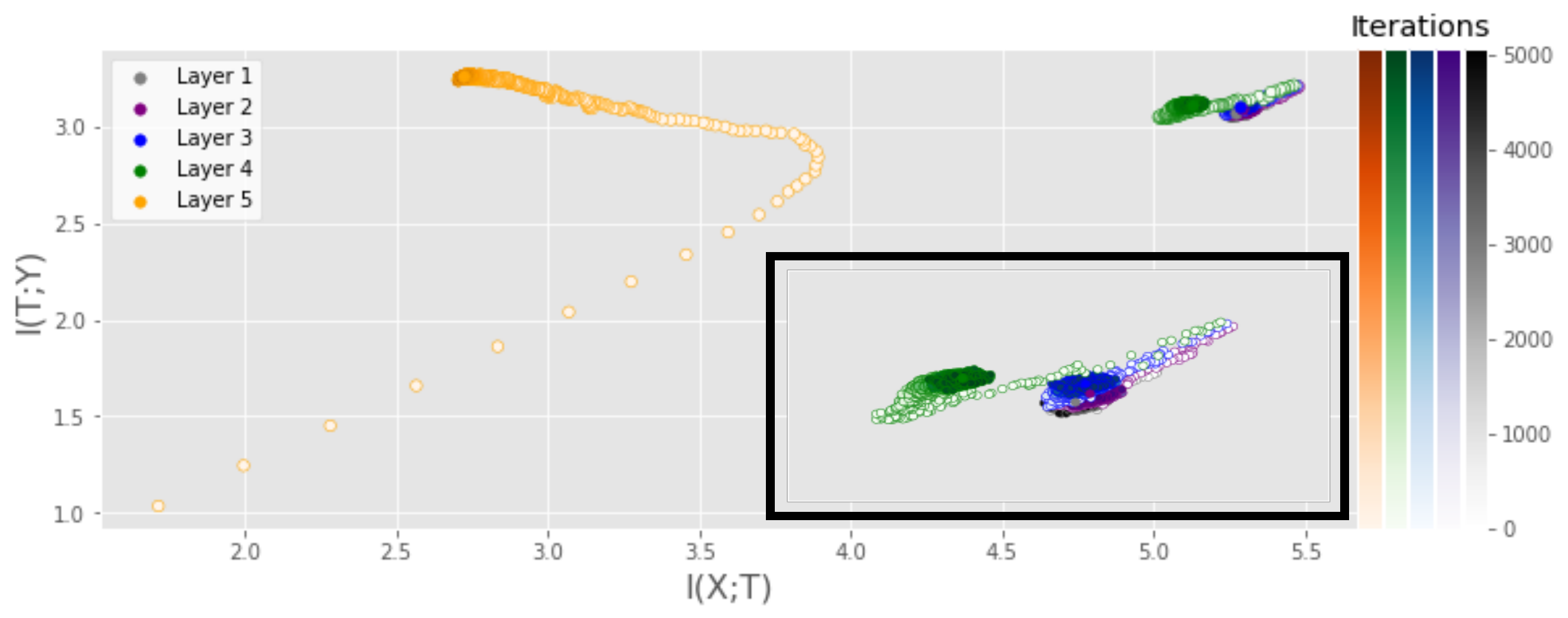}
    \caption{IP of a CNN consisting of three convolutional layers with 4, 8 and 12 filters and one fully connected layer with 256 neurons and a ReLU activation function in in each hidden layer. MI was estimated using the training data of the MNIST dataset and averaged over 5 runs.}
    \label{fig:cnn_relu}
\end{figure*}

\section{Experimental Results}\label{sec:experiments}

We evaluate our approach by comparing it to previous results obtained on small networks by considering the MNIST dataset and an Multi Layer Perceptron (MLP) architecture that was inspired by \cite{michael2018on}. We further compare to a small CNN architecture similar to that of \cite{8683351}, before considering large networks, namely VGG16, and a more challenging dataset, namely CIFAR-10. Note, unless stated otherwise, we use CNN to denote the small CNN architecture.
Details about the MLP and the CNN utilized in these experiments can be found in Appendix \ref{sec:appendixB}. All MI estimates were computed using Equation~\ref{eq:shannon},~\ref{eq:joint renyi entropy}~and~\ref{eq:mutual info} and the tensor approach described in Section 4.

Since the MI is computed at the mini-batch level a certain degree of noise is present. To smooth the MI estimates we employ a moving average approach where each sample is averaged over $k$ mini-batches. For the MLP and CNN experiments we use $k=10$ and for the VGG16 we use $k=50$. We use a batch size of 100 samples, and determine the kernel width using the kernel alignment loss defined in Equation~\ref{eq:kernel loss}. For each hidden layer, we chose the kernel width that maximizes the kernel alignment loss in the range 0.1 and 10 times the mean distance between the samples in one mini-batch. Initially, we sample 75 equally spaced values for the kernel width in the given range for the MLP and CNN and 300 values for the VGG16 network. During training, we dynamically reduce the number of samples to 50 and 100 respectively in to reduce computational complexity and motivated by the fact that the kernel width remains relatively stable during the latter part of training (illustrated in Appendix~\ref{sec:kernelwidth_exp}). We chose the range 0.1 and 10 times the mean distance between the samples in one mini-batch to avoid the kernel width becoming too small and to ensure that we cover a wide enough range of possible values. For the input kernel width we empirically evaluated values in the range 2-16  and found consistent results for values in the range 4-12. All our experiments were conducted with an input kernel width of 8. For the label kernel matrix, we want a kernel width that is as small as possible to approach an ideal kernel matrix, while at the same time large enough to avoid numerical instabilities. For all our experiments we use a value of 0.1 for the kernel width of the label kernel matrix.

\begin{figure*}[tb]
    \centering
    \includegraphics[trim={0 0.2cm 0 0.2cm}, clip, width=\textwidth]{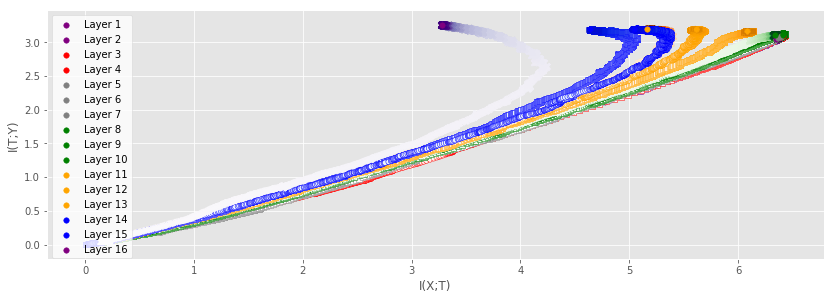}
    \caption{IP of the VGG16 on the CIFAR-10 dataset. MI was estimated using the training data and averaged over 2 runs. Color saturation increases as training progresses. Both the fitting phase and the compression phase is clearly visible for several layers.}
    \label{fig:vgg_train}
\end{figure*}

\begin{figure*}[tb]
    \centering
    \includegraphics[trim={0 0.2cm 0 0.2cm}, clip, width=\textwidth]{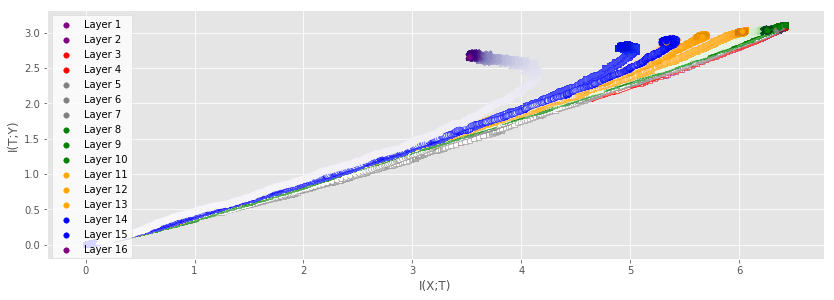}
    \caption{IP of the VGG16 on the CIFAR-10 dataset. MI was estimated using the test data and averaged over 2 runs. Color saturation increases as training progresses. The fitting phase is clearly visible while the compression phase can only be seen in the output layer.}
    \label{fig:vgg_test}
\end{figure*}

{\bf Comparison to Previous Approaches}\hspace{0.2cm} First, we study the IP of the MLP examined in previous works on DNN analysis using information theory~\citep{8683351,michael2018on}. We utilize stochastic gradient descent with a learning rate of 0.09, a cross-entropy loss function, and repeat the experiment 5 times. Figure~\ref{fig:front_page} displays the IP of the MLP with a ReLU activation function in each hidden layer. MI was estimated using the training data of the MNIST dataset. A similar experiment was performed with the tanh activation function, obtaining similar results. The interested reader can find these results in Appendix \ref{sec:appendixC}.

From Figure~\ref{fig:front_page} one can clearly observe a fitting phase, where both $I(T;X)$ and $I(Y;T)$ increases rapidly, followed by a compression phase where $I(T;X)$ decrease and $I(Y;T)$ remains unchanged. Also note that $I(Y;T)$ for the output layer (layer 5 in Figure~\ref{fig:front_page}) stabilizes at an approximate value of $\log_2(10)$, which is to be expected. This can be seen by noting that when the network achieves approximately $100\%$ accuracy, $I(Y;\hat{Y})\approx S(Y)$, where $\hat{Y}$ denotes the output of the network, since $Y$ and $\hat{Y}$ will be approximately identical and the MI between a variable and itself is just the entropy of the variable. The entropy of $Y$ is estimated using Equation~\ref{eq:shannon}, which requires the computation of the eigenvalues of the label kernel matrix $\frac{1}{N} \K_y$.
For the ideal case, where $(\K_y)_{ij} = 1$ if $y_i = y_j$ and zero otherwise,
    $\K_y$ is a rank $K$ matrix, where $K$ is the number of classes in the data.
Thus, $\frac{1}{N} \K_y$ has $K$ non--zero eigenvalues which are given by
    $\lambda_k(\frac{1}{N} \K_y) = \frac{1}{N} \lambda_k(\K_y) = \frac{N_{c_k}}{N}$,
    where $N_{c_k}$ is the number of datapoints in class $k,\; k = 1, 2, \ldots, K$.
Furthermore, if the dataset is balanced we have $N_{c_1} = N_{c_2} = \ldots = N_{c_K} \equiv N_c$.
Then, $\lambda_k\left(\frac{1}{N} \K_y \right) = \frac{N_c}{N} = \frac{1}{K}$, which gives us the entropy estimate
\begin{equation}
    \begin{aligned}
       S\left(\frac{1}{N} \K_y\right) &=
       -\sum\limits_{k=1}^K\lambda_k\left(\frac{1}{N} \K_y\right)\log_2\left[\lambda_k\left(\frac{1}{N} \K_y\right)\right] \\
       &= -\sum_{k = 1}^K \frac{1}{K} \log_2\left[ \frac{1}{K}\right] = \log_2[K].
    \end{aligned}
\end{equation}

Next we examine the IP of a CNN, similar to that studied by \cite{8683351}, with a similar experimental setup as for the MLP experiment. Figure~\ref{fig:cnn_relu} displays the IP of the CNN with a ReLU activation function in all hidden layers. A similar experiment was conducted using the tanh activation function and can be found in Appendix \ref{sec:appendixD}. While the output layer behaves similarly to that of the MLP, the preceding layers show much less movement. In particular, no fitting phase is observed, which might be a result of the convolutional layers being able to extract the necessary information in very few iterations. Note that the output layer is again settling at the expected value $\log_2(10)$, similar to the MLP, as it also achieves close to $100\%$ accuracy.

{\bf Increasing DNN size}\hspace{0.2cm} Finally, we analyze the IP of the VGG16 network on the CIFAR-10 dataset, with the same experimental setup as in the previous experiments. To our knowledge, this is the first time that the full IP has been modeled for such a large-scale network. Figure~\ref{fig:vgg_train} and \ref{fig:vgg_test} show the IP when estimating the MI for the training dataset and the test dataset respectively. For the training dataset, we can clearly observe the same trend as for the smaller networks, where layers experience a fitting phase during the early stages of training and a compression phase in the later stage. Note, that the compression phase is less prominent for the testing dataset. Also note the difference between the final values of $I(Y;T)$ for the output layer estimated using the training and test data, which is a result of the different accuracy achieved on the training data ($\approx 100\%$) and test data ($\approx 90\%)$. \cite{Cheng_2018_ECCV} claim that $I(T;X)\approx H(X)$ and $I(Y;T)\approx H(Y)$ for high dimensional data, and highlight particular difficulties with estimating the MI between convolutional layers and the input/output. However, this statement is dependent on their particular estimator for the MI, and the results presented in Figure~\ref{fig:vgg_train}~and~\ref{fig:vgg_test} demonstrate that neither $I(T;X)$ nor $I(Y;T)$ is deterministic for our proposed estimator.

{\bf Effect of Early Stopping}\hspace{0.2cm} We also investigate the effect of using early stopping on the IP described above. Early stopping is a regularization technique where the validation accuracy is monitored and training is stopped if the validation accuracy does not increase for a set number of iteration, often referred to as the patience hyperparameter. Figure~\ref{fig:front_page} displays the results of monitoring where the training would stop if the early stopping procedure was applied for different values of patience. For a patience of 5 iterations the network training would stop before the compression phase takes place for several of the layers. For larger patience values, the effects of the compression phase can be observed before training is stopped. Early stopping is a procedure intended to prevent the network from overfitting, which may imply that the compression phase observed in the IP of DNNs can be related to overfitting.

{\bf Data Processing Inequality}\hspace{0.2cm} A DNN consists of a chain of mappings from the input, through the hidden layers and to the output. One can interpret a DNN as a Markov chain \citep{DBLP:journals/corr/Shwartz-ZivT17, DBLP:journals/corr/abs-1804-00057} that defines an information path \citep{DBLP:journals/corr/Shwartz-ZivT17}, which should satisfy the following Data Processing Inequality \citep{Cover}:
\begin{equation}\label{eq:dpi} I(X;T_1) \geq I(X;T_2) \geq \ldots \geq I(X;T_L),
\end{equation} 
where $L$ is the number of layers in the network. An indication of a good MI estimator is that it tends to uphold the DPI. Figure~\ref{fig:dpi_plot} in Appendix \ref{sec:dpi} illustrates the mean difference in MI between two subsequent layers in the MLP and VGG16 network. Positive numbers indicate that MI decreases, thus indicating compliance with the DPI. We observe that our estimator complies with the DPI for all layers in the MLP and all except one in the VGG16 network.

\section{Conclusion}

In this work, we propose a novel framework for analyzing DNNs from a MI perspective using a tensor-based estimate of the R\'enyi's $\alpha$-Order Entropy. Our experiments illustrate that the proposed approach scales to large DNNs, which allows us to provide insights into the training dynamics. We observe that the compression phase in neural network training tends to be more prominent when MI is estimated on the training set and that commonly used early-stopping criteria tend to stop training before or at the onset of the compression phase. This could imply that the compression phase is linked to overfitting. Furthermore, we showed that, for our tensor-based approach, the claim that $H(X) \approx I(T;X)$ and $H(Y) \approx I(T;Y)$ does not hold. We believe that our proposed approach can provide new insight and facilitate a more theoretical understanding of DNNs.

\bibliography{iclr2020_conference}
\bibliographystyle{iclr2020_conference}

\appendix

\section{Bound on Matrix--Based Entropy Estimator}\label{sec:bound}

Not all estimators of entropy have the same properties. The ones developed for Shannon suffer from the curse of dimensionality \citep{1114861}. In contrast, Renyi's entropy estimators have the same functional form of the statistical quantity in a Reproducing Kernel Hilbert Space (RKHS), thus capturing properties of the data population. Essentially, we are projecting marginal distribution to a RKHS in order to measure entropy and MI. This is similar to the approach of maximum mean discrepancy and the kernel mean embedding \citep{Gretton:2012:KTT:2188385.2188410, fuku}. The connection with the data population can be shown via the theory of covariance operators. The covariance operator $G: \mathcal{H} \rightarrow \mathcal{H}$ is defined through the bilinear form  

\begin{equation*}
\mathcal{G}(f,g) = \left< f, Gg \right> = \int_\mathcal{X} \left< f, \psi(\x) \right> \left< \psi(\x), g \right> d \mathbb{P}_\mathcal{X}(\x) =  E_\mathcal{X} \left\{ f(X),g(Y) \right\}
\end{equation*}

where $\mathbb{P}_\mathcal{X}$ is a probability measure and $f,g \in \mathcal{H}$. Based on the empirical distribution $\mathbb{P}_N = \frac{1}{N} \sum_{i=1}^N \delta_{\x_i} (\x)$, the empirical version $\hat{G}$ of $G$ obtained from a sample $\x_i$ of size N is given by:

\begin{equation*}
\left<f,\hat{G}_{Ng}\right> = \mathcal{\hat{G}}(f,g) = \int_\mathcal{X} \left< f, \psi(\x) \right> \left< \psi(\x), g \right> d \mathbb{P}_\mathcal{X}(\x) =  \frac{1}{N}\sum_{i=1}^N\left<f, \psi(\x_i)\right>\left<\psi(\x_i), g\right>
\end{equation*}

By analyzing the spectrum of $\hat{G}$ and $G$, \cite{DBLP:journals/corr/abs-1211-2459} showed the the different between $\tr(G)$ and $\tr(\hat{G})$ can be bounded, as stated in the following proposition:

\begin{proposition}\label{prop:bound}
Let $\mathbb{P}_N = \frac{1}{N} \sum_{i=1}^N \delta_{\x_i} (\x)$ be the empirical distribution. Then, as a consequence of Proposition 6.1 in \cite{DBLP:journals/corr/abs-1211-2459}, $\mbox{tr}\left[ \hat{G}^\alpha_N \right] = \mbox{tr}\left[ \left( \frac{1}{N} \K \right)^\alpha \right]$. The difference between $\tr(G)$ and $\tr(\hat{G})$ can be bounded under the conditions of Theorem 6.2 in \cite{DBLP:journals/corr/abs-1211-2459} and for $\alpha>1$, with probability 1-$\delta$

\begin{equation}\label{eq:bound}
\left| \mbox{tr}\left( G^\alpha \right) - \mbox{tr}\left( \hat{G}^\alpha_N \right) \right| \leq \alpha C \sqrt{\frac{2 \log \frac{2}{\delta}}{N}}
\end{equation}
where C is a compact self-adjoint operator.
\end{proposition}

\section{Proof of Equation \ref{eq:shannon} in Section \ref{sec:matrix_renyi}} \label{sec:appendixA}

\begin{proof}
\begin{equation*}
    \begin{aligned}
        \lim_{\alpha \rightarrow 1} S_\alpha(\mathbf A)
            &= \lim_{\alpha \rightarrow 1} \frac{1}{1 - \alpha} \log_2\left( \sum_{i = 1}^n \lambda_i^\alpha \right) \rightarrow \frac{0}{0},
    \end{aligned}
\end{equation*}
    since $\sum_{i = 1}^N \lambda_i = \tr(\mathbf A) = 1$.
L'H\^{o}pital's rule yields
\begin{equation}
    \begin{aligned}
    \lim_{\alpha \rightarrow 1} S_\alpha(\mathbf A)
        &= \lim_{\alpha \rightarrow 1}
                \frac{
                    \frac{\partial}{\partial \alpha} \log_2\left[\sum_{i = 1}^n \lambda_i(\mathbf A)^\alpha \right]
                }{
                    \frac{\partial}{\partial \alpha} (1 - \alpha)  
                } \\
        &= -\frac{1}{\ln 2} \lim_{\alpha \rightarrow 1}
                \frac{
                    \sum_{i = 1}^n \lambda_i(\mathbf A)^\alpha \ln[\lambda_i(\mathbf A)] 
                }{
                    |\sum_{i = 1}^n \lambda_i(\mathbf A)^\alpha|
                } \\
        &= -\sum_{i = 1}^n \lambda_i(\mathbf A) \log_2[\lambda_i(\mathbf A)].
    \end{aligned}
\end{equation}
\end{proof}

\section{Detailed Description of Networks from Section \ref{sec:experiments}} \label{sec:appendixB}

We provide a detailed description of the architectures utilized in Section \ref{sec:experiments} of the main paper. Weights were initialized according to \cite{He2015DelvingDI} when the ReLU activation function was applied and initialized according to~\cite{pmlr-v9-glorot10a} for the experiments conducted using the tanh activation function. Biases were initialized as zeros for all networks. All networks were implemented using the deep learning framework Pytorch \citep{paszke2017automatic}.

\subsection{Multilayer Perceptron Utilized in Section \ref{sec:experiments}}

The MLP architecture used in our experiments is the same architecture utilized in previous studies on the IP of DNN \citep{8683351,michael2018on}, but with Batch Normalization \citep{Ioffe2015BatchNA} included after the activation function of each hidden layer. Specifically, the MLP in Section 5 includes (from input to output) the following components: 

\begin{enumerate}
    \item Fully connected layer with 784 inputs and 1024 outputs.
    \item Activation function.
    \item Batch normalization layer
    \item Fully connected layer with 1024 inputs and 20 outputs.
    \item Activation function.
    \item Batch normalization layer
    \item Fully connected layer with 20 inputs and 20 outputs.
    \item Activation function.
    \item Batch normalization layer
    \item Fully connected layer with 20 inputs and 20 outputs.
    \item Activation function.
    \item Batch normalization layer
    \item Fully connected layer with 784 inputs and 10 outputs.
    \item Softmax activation function.
\end{enumerate}

\subsection{Convolutional Neural Network Utilized in Section \ref{sec:experiments}}

The CNN architecture in our experiments is a similar architecture as the one used by \cite{8683351}. Specifically, the CNN in Section 5 includes (from input to output) the following components: 

\begin{enumerate}
    \item Convolutional layer with 1 input channel and 4 filters, filter size $3\times 3$, stride of 1 and no padding.
    \item Activation function.
    \item Batch normalization layer
    \item Convolutional layer with 4 input channels and 8 filters, filter size $3\times 3$, stride of 1 and no padding.
    \item Activation function.
    \item Batch normalization layer
    \item Max pooling layer with filter size $2 \times 2$, stride of 2 and no padding.
    \item Convolutional layer with 8 input channels and 16 filters, filter size $3\times 3$, stride of 1 and no padding.
    \item Activation function.
    \item Batch normalization layer
    \item Max pooling layer with filter size $2 \times 2$, stride of 2 and no padding.
    \item Fully connected layer with 400 inputs and 256 outputs.
    \item Activation function.
    \item Batch normalization layer
    \item Fully connected layer with 256 inputs and 10 outputs.
    \item Softmax activation function.
\end{enumerate}

\section{IP of MLP with tanh activation function from Section \ref{sec:experiments}}\label{sec:appendixC}

\begin{figure*}[tb]
    \centering
    \includegraphics[width=\linewidth]{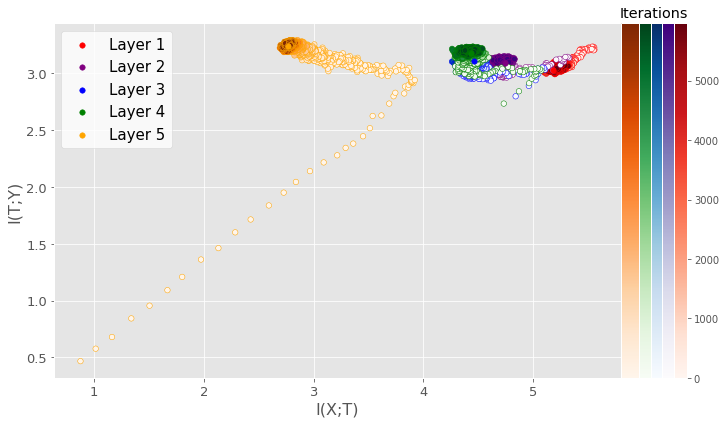}
    \caption{IP of a MLP consisting of four fully connected layers with 1024, 20, 20, and 20 neurons and a tanh activation function in in each hidden layer. MI was estimated using the training data of the MNIST dataset and averaged over 5 runs.}
    \label{fig:mlp_tanh}
\end{figure*}

Figure~\ref{fig:mlp_tanh} displays the IP of the MLP described above with a tanh activation function applied in each hidden layer. Similarly to the ReLU experiment in the main paper, a fitting phase is observed, where both $I(T;X)$ and $I(Y;T)$ increases rapidly, followed by a compression phase where $I(T;X)$ decrease and $I(Y;T)$ remains unchanged. Also note that, similar to the ReLU experiment, $I(Y;T)$ stabilizes close to the theoretical maximum value of $\log_2(10)$.

\section{IP of CNN with tanh activation function from Section \ref{sec:experiments}}\label{sec:appendixD}

\begin{figure*}[tb]
    \centering
    \includegraphics[width=\linewidth]{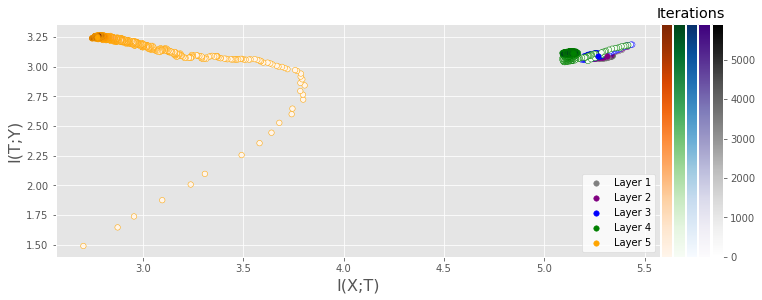}
    \caption{IP of a CNN consisting of three convolutional layers with 4, 8 and 12 filters and one fully connected layer with 256 neurons and a tanh activation function in in each hidden layer. MI was estimated using the training data of the MNIST dataset and averaged over 5 runs.}
    \label{fig:cnn_tanh}
\end{figure*}

Figure~\ref{fig:cnn_tanh} displays the IP of the CNN described above with a tanh activation function applied in each hidden layer. Just as for the CNN experiment with ReLU activation function in the main paper, no fitting phase is observed for the majority of the layers, which might indicate that the convolutional layers can extract the essential information after only a few iterations.

\section{Kernel width sigma}\label{sec:kernelwidth_exp}

\begin{figure*}
    \centering
    \begin{subfigure}{0.32\linewidth}
    \includegraphics[width=\textwidth]{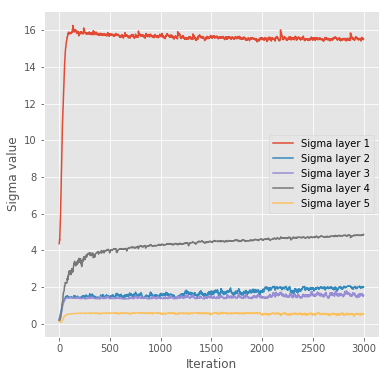}
    \caption{MLP}
    \end{subfigure}
    \begin{subfigure}{0.32\linewidth}
    \includegraphics[width=\textwidth]{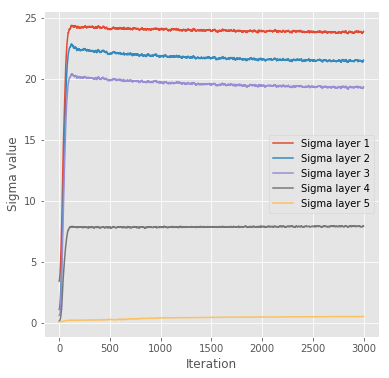}
    \caption{CNN}
    \end{subfigure}
    \begin{subfigure}{0.32\linewidth}
    \includegraphics[width=\textwidth]{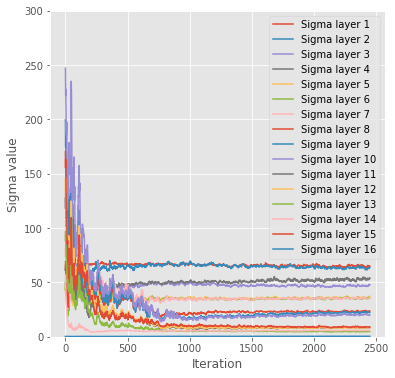}
    \caption{VGG16}
    \end{subfigure}
    \caption{Evolution of kernel width as a function of iteration for the three networks that we considered in this work. The plots demonstrate how the optimal kernel width quickly stabilizes and stay relatively stable throughout the training.}
    \label{fig:sigma_plot}
\end{figure*}

We further evaluate our dynamic approach of finding the kernel width $\sigma$. Figure~\ref{fig:sigma_plot} shows the variation of $\sigma$ in each layer for the MLP, the small CNN and the VGG16 network. We observe that the optimal kernel width for each layer (based on the criterion in Section~\ref{sec:kernel_width}), stabilizes reasonably quickly and remains relatively constant during training. This illustrates that decreasing the sampling range is a meaningful approach to decrease computational complexity.

\section{Data Processing Inequality}\label{sec:dpi}

\begin{figure*}
    \centering
    \begin{subfigure}{0.49\linewidth}
    \includegraphics[width=\textwidth]{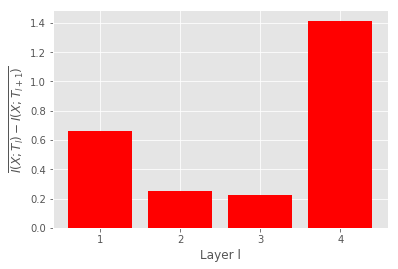}
    \caption{MLP}
    \end{subfigure}
    \begin{subfigure}{0.5\linewidth}
    \includegraphics[width=\textwidth]{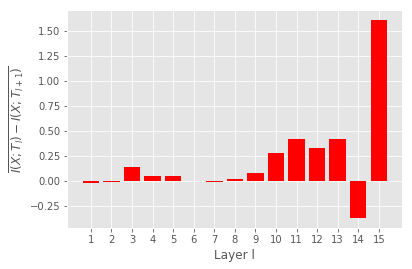}
    \caption{VGG16}
    \end{subfigure}
    \caption{Mean difference in MI of subsequent layers $\ell$ and $\ell+1$. Positive numbers indicate compliance with the DPI. MI was estimated on the MNIST training set for the MLP and on the CIFAR-10 training set for the VGG16.}
    \label{fig:dpi_plot}
\end{figure*}

Figure~\ref{fig:dpi_plot} illustrates the mean difference in MI between two subsequent layers in the MLP and VGG16 network. Positive numbers indicate that MI decreases, thus indicating compliance with the DPI. We observe that our estimator complies with the DPI for all layers in the MLP and for all except one in the VGG16 network.

Note, the difference in MI is considerably lower for the early layers in the network, which is further shown by the grouping of the early layers for our convolutional based architectures (Figure~\ref{fig:cnn_relu}-\ref{fig:vgg_test}).

\end{document}